\documentclass[conference]{IEEEtran}
\IEEEoverridecommandlockouts

\usepackage{cite}
\usepackage{amsmath,amssymb,amsfonts}
\usepackage{algorithmic}
\usepackage{graphicx}
\usepackage{textcomp}
\usepackage{xcolor}
\usepackage{flushend}

\usepackage{subcaption}
\usepackage{booktabs, multicol}
\usepackage{url}

\def\BibTeX{{\rm B\kern-.05em{\sc i\kern-.025em b}\kern-.08em
    T\kern-.1667em\lower.7ex\hbox{E}\kern-.125emX}}

\begin{document}

\title{Multi-Scale Denoising in the Feature Space for Low-Light Instance Segmentation\\

\thanks{This work was supported by the UKRI MyWorld Strength in Places Programme (SIPF00006/1) and EPSRC Doctoral Training Partnerships (EP/W524414/1)}
}

\author{\IEEEauthorblockN{Joanne Lin}
\IEEEauthorblockA{\textit{Visual Information Laboratory} \\
\textit{University of Bristol}\\
Bristol, United Kingdom \\
joanne.lin@bristol.ac.uk}
\and
\IEEEauthorblockN{Nantheera Anantrasirichai}
\IEEEauthorblockA{\textit{Visual Information Laboratory} \\
\textit{University of Bristol}\\
Bristol, United Kingdom \\
n.anantrasirichai@bristol.ac.uk}
\and
\IEEEauthorblockN{David Bull}
\IEEEauthorblockA{\textit{Visual Information Laboratory} \\
\textit{University of Bristol}\\
Bristol, United Kingdom \\
dave.bull@bristol.ac.uk}
}

\maketitle

\begin{abstract}
Instance segmentation for low-light imagery remains largely unexplored due to the challenges imposed by such conditions, for example shot noise due to low photon count, color distortions and reduced contrast. In this paper, we propose an end-to-end solution to address this challenging task.
Our proposed method implements weighted non-local blocks (wNLB) in the feature extractor. This integration enables an inherent denoising process at the feature level. As a result, our method eliminates the need for aligned ground truth images during training, thus supporting training on real-world low-light datasets. We introduce additional learnable weights at each layer in order to enhance the network's adaptability to real-world noise characteristics, which affect different feature scales in different ways.
Experimental results on several object detectors show that the proposed method outperforms the pre-trained networks with an Average Precision (AP) improvement of at least \textbf{+7.6}, with the introduction of wNLB further enhancing AP by upto \textbf{+1.3}.
\end{abstract}


\section{Introduction}
\label{sec:intro}

Instance segmentation is a crucial task in computer vision because it allows systems to interpret scenes with higher levels of detail, by understanding the position and shape of the objects within them. This can be highly beneficial in scenarios such as autonomous driving, video surveillance, multimedia editing, and agricultural monitoring. Its importance is underscored by numerous recent research contributions in this field and various benchmarks and datasets \cite{lin2014coco, zhou2017ade20k, Everingham15pascalvoc, cordts2016cityscapes}. 

\begin{figure}[t]
    \centering
    \includegraphics[width=\linewidth]{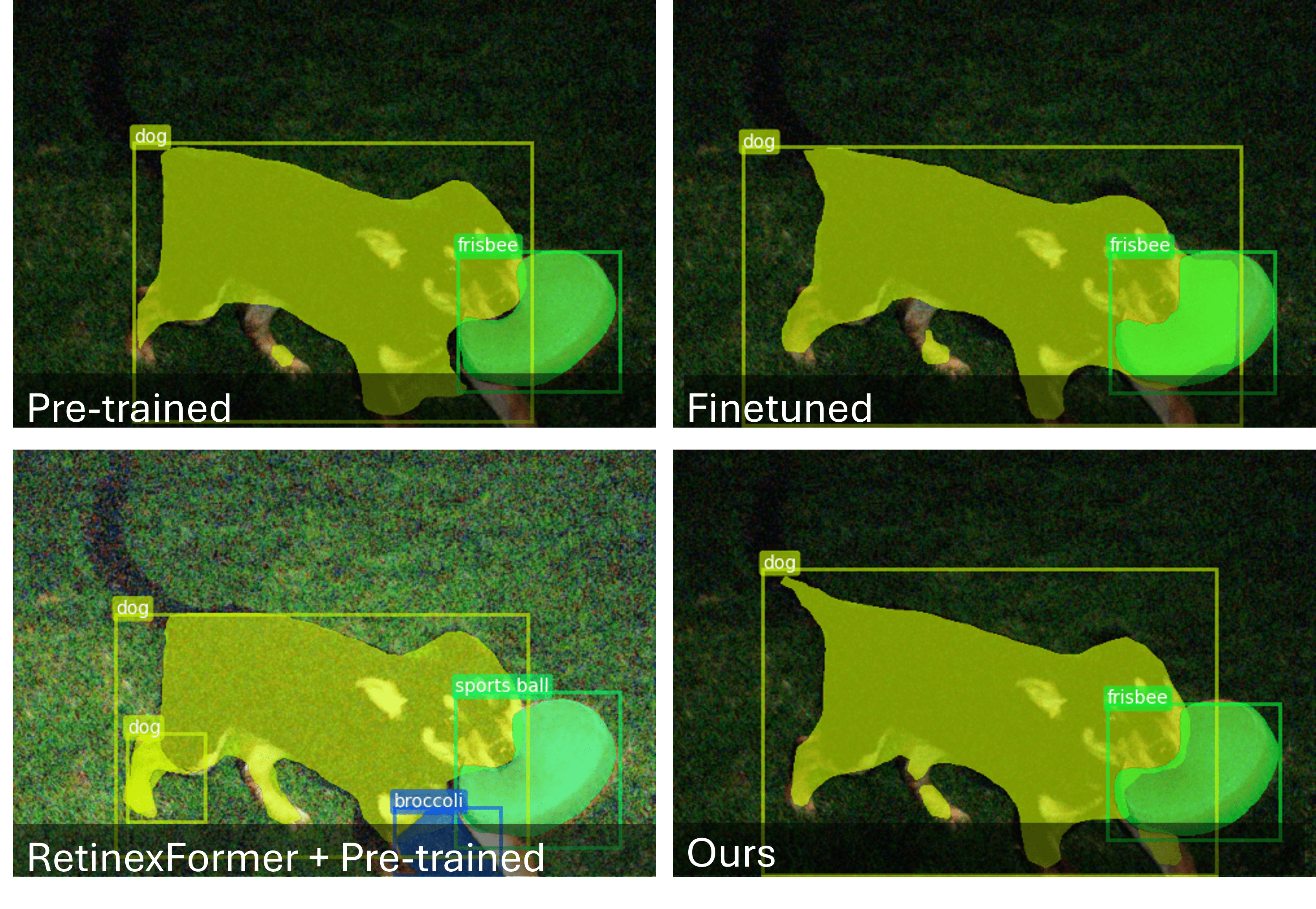}
    \caption{Visual comparison of low-light instance segmentation for different methods using Mask R-CNN \cite{he2017maskrcnn} as the detector.}
    \label{fig:intro_comparison}
    \vspace{-1em}
\end{figure}

While its applications span many sectors, there has been limited research on instance segmentation in low-light conditions. This is because low-light images are heavily affected by noise from a low photon count in darker conditions, making the task much more challenging. A common approach is to first apply a low-light image enhancement (LLIE) method (such as \cite{jiang2021enlightengan, li2021zerodce, lv2021agllnet, cai2023retinexformer}) to these images before passing into existing instance segmentation methods.
Specifically, Zero-DCE \cite{li2021zerodce}, and RetinexFormer \cite{cai2023retinexformer} also use low-light object detection benchmarks (i.e. \cite{2018exdark, 2020poorvisdata}) to demonstrate their enhancing performances. To the best of our knowledge to date, the only known work on low-light instance segmentation is by Chen et al. \cite{2023lis}, focusing on leveraging the higher bit-depth in RAW images for better performance.

Our goal is to develop a robust and effective end-to-end method for low-light instance segmentation. This implies eliminating any explicit pre-processing steps in order to simplify the mask generating procedure. However, most models are not designed to handle low-light images; in our experiments, we found that both low-contrast and noise impact the performance of these models, as also confirmed by the study in \cite{Yi:Comprehensive:2024}.

In this paper, we propose integrating ``plug-and-play'' denoising blocks into existing feature extractors, for an end-to-end approach to low-light instance segmentation. Our proposed weighted non-local blocks (wNLB) are an extension to non-local blocks (NLB) \cite{wang2018nonlocalnn}, where the additional learnable weight ensures the network can learn to ignore appropriate noise characteristics at different feature scales.
We trained our framework on a synthetic dataset due to the infancy of machine perception tasks in low-light conditions, focusing on sRGB images as they are more widely used than RAW images and are also applicable to video sequences. We used the popular Microsoft COCO \cite{lin2014coco} dataset and low-light simulation pipeline proposed by \cite{lv2021agllnet}.
Our comprehensive experiments on both synthetic and real low-light datasets show our wNLB can improve existing networks performances by upto 1.3AP.

\section{Related Work}
\label{sec:related}

\subsection{Instance Segmentation}
\label{ssec:inseg}
Instance segmentation aims to produce a segmentation mask, bounding box and class label for every object detected in an image. One of the most popular techniques for achieving this is Mask R-CNN \cite{he2017maskrcnn}, introduced by He et al. as an extension to the popular object detection model, Faster R-CNN \cite{ren2015fasterrcnn}.
With the notable work from \cite{vaswani2017attention}, there has been a surge in transformer-based methods in recent years. For example, DETR-based \cite{carion2020detr} instance segmentation methods use CNN-based feature extraction backbones and concatenate positional embeddings before passing into a transformer. However, these often have many parameters and require a lot of computational resources to train effectively.
One-stage methods \cite{jocher2023yolo, wang2020solov2} aim to remove the efficiency problem by skipping the region proposal stage found in two-stage methods like Mask R-CNN. With less parameters and less computation required, they are often used in real-time situations, e.g. autonomous driving.

\subsection{Low-light image enhancement}
\label{ssec:llie}
Traditional LLIE methods treat the problem as two separate tasks: contrast enhancement and denoising. Contrast enhancement methods include gamma correction and adaptive histogram equalizatiom (AHE) \cite{pizer1987histeq}; however these lack consideration of illumination factors, leading to cognition approaches inspired by Retinex Theory \cite{land1977retinex}.
Contrast-enhancement of low-light images, however, amplifies noise which necessitates the use of denoising algorithms for further refinement.

Recent advancements in LLIE have explored the use of deep learning. Lore et al. first introduced LLNet \cite{lore2017llnet}, an autoencoder trained with synthetic low-light images.
Supervised methods, including RetinexNet \cite{Chen2018RetinexNet}, KiND \cite{zhang2019kindling}, and AGLLNet \cite{lv2021agllnet}, require large sets of paired normal/low-light images for training, often generated synthetically.
Unsupervised methods gained popularity due to the scarcity of paired normal-light images in low-light datasets. Examples include Zero-DCE (and its extension Zero-DCE++) \cite{li2021zerodce}. RetinexFormer \cite{cai2023retinexformer}, a transformer-based method, is currently the state-of-the-art approach, reformulating the Retinex theory \cite{land1977retinex} to account for noise as well when decomposing images into reflectance and luminance components.

\subsection{Low-light scene understanding}
\label{ssec:llsu}
Low-light scene understanding is an under-explored field due to the need for high-quality annotations.
Datasets such as NightCity \cite{2021nightcity}, BDD100K-night \cite{2020bdd100k}, ExDark \cite{2018exdark} and \cite{2020poorvisdata} are datasets that have been released for several low-light scene understanding tasks, however these do not tackle the instance segmentation problem.
As far as we are aware, Chen et al. \cite{2023lis} have released the only low-light instance segmentation dataset to date. However, they only provide 2230 images, which is not suitable for training. Furthermore, their research focused on leveraging RAW data for \textit{extreme} low-light instance segmentation.

\section{Proposed Method}
\label{sec:method}

\begin{figure}[!t]
  \centering
  \includegraphics[width=\linewidth]{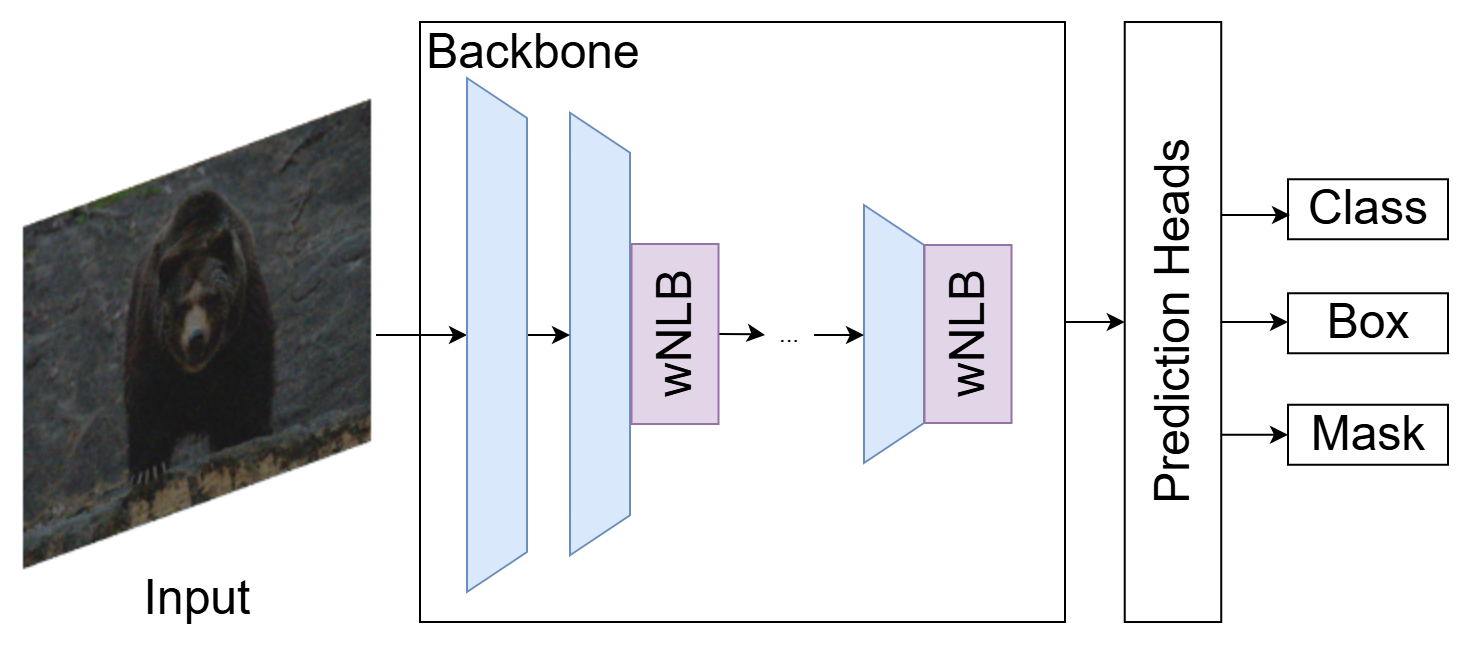}
\caption{Generic architecture showing our proposed weighted non-local blocks added into the backbone to remove noise in the feature space. Blue blocks indicate convolutional layers.}
\label{fig:architecture}
\end{figure}

\begin{figure}[!t]
    \centering
    \includegraphics[width=\linewidth]{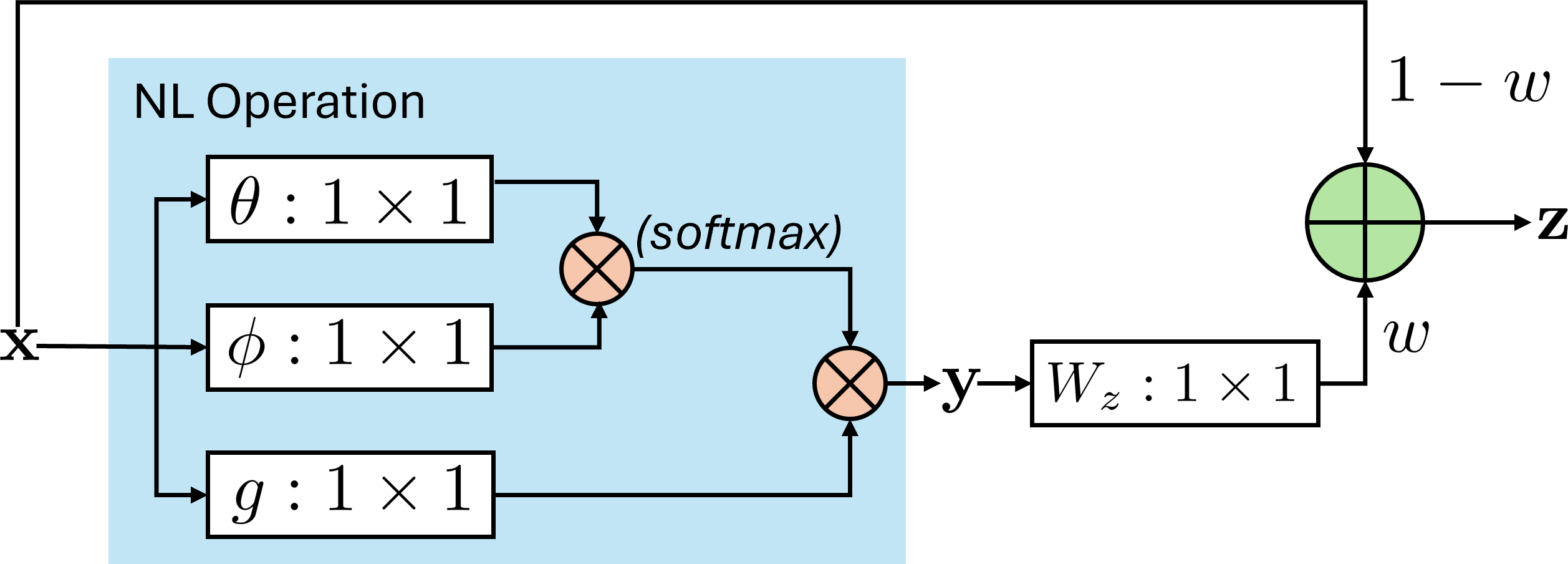}
    \caption{Our proposed weighted non-local block (wNLB) for feature denoising with learnable weight $w$.
    }
    \label{fig:nonlocal_block_diagram}
    \vspace{-1em}
\end{figure}

Non-local means (NLM) \cite{buades2005nlm} is a class of traditional computer vision denoising algorithms that process images on a global scale, unlike local smoothing filters (e.g. bilateral filter \cite{tomasi1998bilateral}, Gaussian filter etc.) which only consider neighbouring pixels. Non-local means filtering involves calculating the mean of all pixels in the image, with the weighting determined by the similarity of each pixel to the target one. Wang et al. \cite{wang2018nonlocalnn} applied this idea in their neural networks to learn long-range dependencies between pixels spatially and temporally. In their experiments, they found that the self-attention mechanism \cite{vaswani2017attention} is a special form of NL operations in the embedded gaussian form \cite{wang2018nonlocalnn}. 

Xie et al. \cite{xie2019nlmdenoise} explored using both non-local means algorithms and local smoothing filtering algorithms for feature denoising to defend against adversarial attacks; attacks that intentionally inject noise into images to confuse computer systems.
They found that although the local denoising operations also improved the robustness of these systems, NL operations were more effective in reducing feature noise.

NLBs consist of an implementation of the non-local means algorithm (e.g. dot product, concatenation, Gaussian, embedded Gaussian) followed by a 1$\times$1 convolution, added to the block's input via an identity skip connection. They achieve feature denoising as they pass the feature maps through an NLM operation.

Our proposed weighted NLBs extend the NLBs to include a learnable weight $w$ as shown in Fig. \ref{fig:nonlocal_block_diagram}. This allows the model to control the amount of denoising at different scales in the feature extractor.
Our wNLBs computes the following:
\begin{equation}
    \mathbf{z} = wW_z\mathbf{y} + (1-w)\mathbf{x}
\end{equation}
\noindent where $\mathbf{x}$ is the input feature map, $\mathbf{y}$ is the output from the NL means operation, $W_z$ is the weight matrix from the 1$\times$1 convolutional layer after the NL means operation and $\mathbf{z}$ is the output.

\begin{figure}[!t]
\centering
\includegraphics[width=\linewidth]{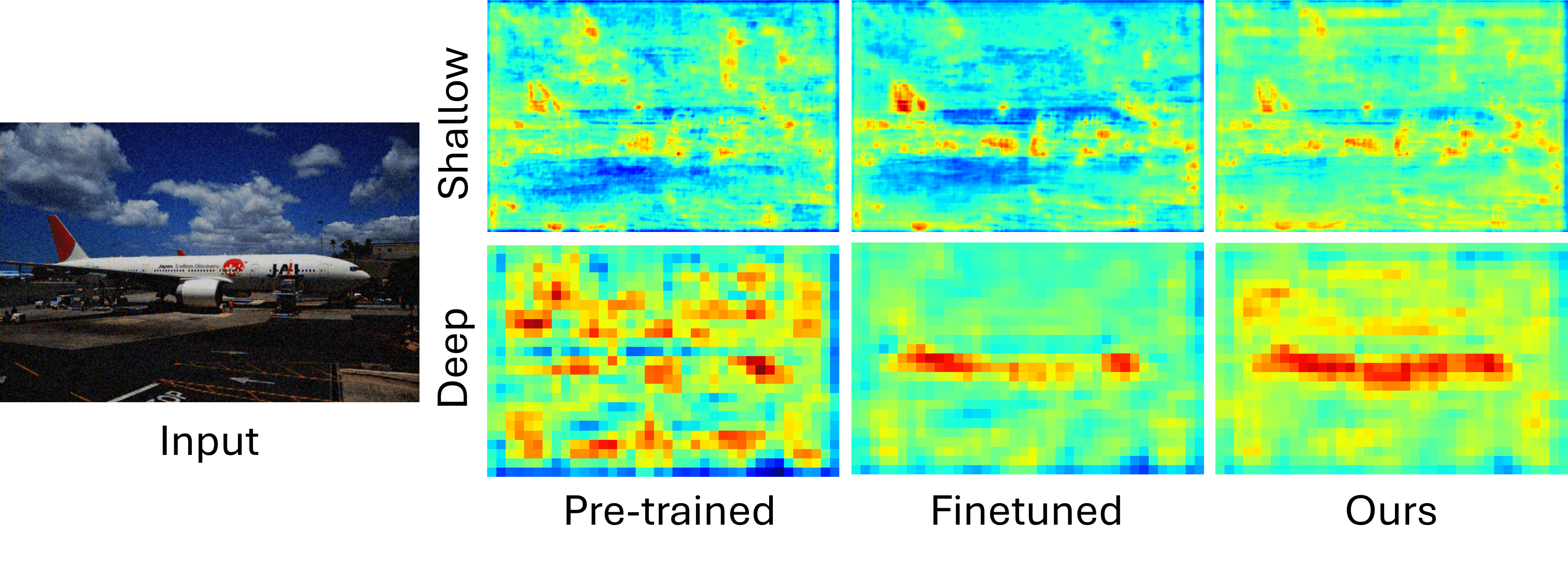}
\caption{Comparison of shallow and deep features extracted from our proposed method against pre-trained and finetuned Mask R-CNN \cite{he2017maskrcnn} models.}
\label{fig:feature_map_comparison}
\end{figure}

\begin{figure}[!t]
\centering
\includegraphics[width=\linewidth]{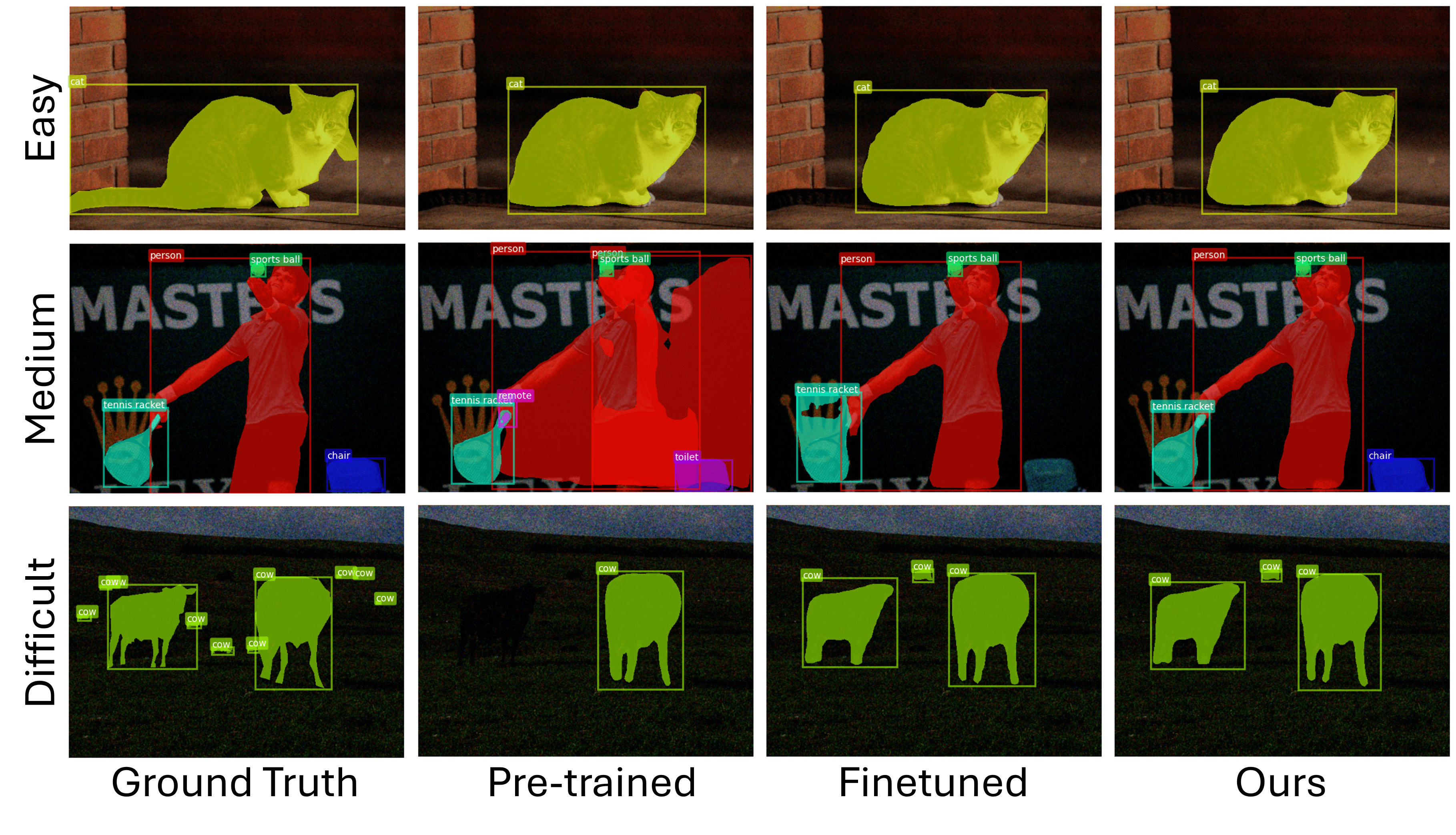}
\caption{Visual comparison of our proposed method against pre-trained and finetuned Mask R-CNN \cite{he2017maskrcnn} models, along with the ground truth, for cases of varying levels of difficulty. 
}
\label{fig:baseline_visual_comparison}
\end{figure}

\begin{table}[!t]
    \caption{Comparison of instance segmentation methods on the synthetic low-light COCO minival dataset}
    \centering
    \resizebox{1.0\linewidth}{!}{\begin{tabular}{llccccccc}
        \toprule
        Method     &  & AP  & AP$_{50}$ & AP$_{75}$ & AP$_{S}$ & AP$_{M}$ & AP$_{L}$ \\
        \midrule
        Mask R-CNN & Pre-trained & 6.9 & 12.4 & 6.9 & 2.3 & 7.4 & 12.4\\
        Mask R-CNN & Finetuned & 15.9 & 28.6 & 15.6 & 4.8 & 15.8 & 27.8 \\
        Mask R-CNN & NLB & 16.6 & 30.3 & 16.4 & 5.4 & 16.7 & 28.1 \\
        Mask R-CNN & wNLB &  \textbf{16.9} & \textbf{30.7} & \textbf{16.6} & \textbf{5.6} & \textbf{17.2} & \textbf{28.9} \\
        \midrule
        YOLOv8 & Pre-trained & 6.3 & 11.1 & 6.3 & 1.8 & 6.8 & 10.7 \\
        YOLOv8 & Finetuned & 14.3 & 25.6 & 14.3 & 4.0 & 14.0 & 24.1 \\
        YOLOv8 & NLB & \textbf{22.1} & \textbf{37.6} & \textbf{22.2} & \textbf{7.4} & 23.1 & \textbf{36.6} \\
        YOLOv8 & wNLB & 22.0 & 37.5 & 22.1 & \textbf{7.4} & \textbf{23.2} & 36.5 \\
        \midrule
        SOLOv2 & Pre-trained & 8.2 & 14.1 & 8.3 & 2.7 & 8.6 & 14.1 \\
        SOLOv2 & Finetuned & 15.0 & 26.5 & 14.9 & 3.9 & 15.0 & 26.5 \\
        SOLOv2 & NLB & \textbf{15.8} & \textbf{27.9} & 15.6 & \textbf{4.1} & \textbf{15.8} & 27.6 \\
        SOLOv2 & wNLB & \textbf{15.8} & \textbf{27.9} & \textbf{15.8} & \textbf{4.1} & 15.7 & \textbf{27.8} \\
        \bottomrule
    \end{tabular}}
    \label{tab:methods_comparison}
    \vspace{-1em}
\end{table}

\begin{table}[!t]
    \caption{Comparison of two-stage the synthetic low-light COCO minival dataset}
    \centering
    \begin{tabular}{lcccccc}
        \toprule
        Method     & AP  & AP$_{50}$ & AP$_{75}$ & AP$_{S}$ & AP$_{M}$ & AP$_{L}$  \\
        \midrule
        EnlightenGAN \cite{jiang2021enlightengan} & 5.5 & 10.0 & 5.5 & 1.7 & 6.1 & 10.0 \\
        ZeroDCE++ \cite{li2021zerodce} & 5.6 & 10.0 & 5.6 & 1.8 & 6.1 & 9.6 \\
        AGLLNet \cite{lv2021agllnet} & 6.1 & 11.0 & 6.1 & 1.8 & 7.2 & 10.5 \\
        RetinexFormer \cite{cai2023retinexformer} & 5.7 & 10.3 & 5.7 & 1.8 & 6.5 & 9.9 \\
        Ours & \textbf{16.9} & \textbf{30.7} & \textbf{16.6} & \textbf{5.6} & \textbf{17.2} & \textbf{28.9} \\
        \bottomrule
    \end{tabular}
    \label{tab:enhance_methods_comparison}
\end{table}

\section{Experimental Results}
\label{sec:experiments}
For all of our experiments we finetuned the models with an initial learning rate of $5 \times 10^{-4}$ for 10 epochs. We include the performances from pre-trained models to demonstrate that models trained on popular datasets are not sufficient for low-light.
To validate the effectiveness of our proposed wNLB, we used 3 different instance segmentation methods: Mask R-CNN \cite{he2017maskrcnn}, YOLOv8 \cite{jocher2023yolo} and SOLOv2 \cite{wang2020solov2}. We used a ResNet-50 FPN \cite{he2016resnet, lin2017fpn} backbone for Mask R-CNN and SOLOv2, and a medium-sized CSPDarkNet \cite{jocher2023yolo} backkone for YOLOv8 as they are of similar sizes to ensure a fair comparison.
Due to limited computational resources (RTX3090), we were unable to add wNLBs at the end of the first layer of the backbones. 

We used the Microsoft COCO \cite{lin2014coco} dataset and passed the training and validation sets through the low-light synthetic pipeline proposed by Lv et al. \cite{lv2021agllnet} to create our dataset for training and evaluation.
We used their validation set for evaluation as their test set annotations are not publicly available.
We also used the metrics provided by MS COCO: 
AP calculates the mean of the average precisions across multiple IoU thresholds, 0.5 to 0.95, in 0.05 increments. AP$_{50}$ and AP$_{75}$ calculate the average precision with an IoU threshold of 0.5 and 0.75 respectively. AP$_S$, AP$_M$, and AP$_L$ calculate the average precision for small, medium and large objects, which are pre-defined by \cite{lin2014coco}.

We experimented with one-stage methods (end-to-end frameworks) and two-stage methods (low-light enhancement followed by instance segmentation). 

\subsection{One-stage experiments}
\label{ssec:baseline_experiments}
Our baseline experiments compare our proposed method against pre-trained models, both before and after fine-tuning them on synthetic low-light training data as described above. Table \ref{tab:methods_comparison} shows a significant improvement between the fine-tuned model and the pre-trained model. Additionally, there is further enhancement when the wNLBs are integrated, resulting in an improvement of at least \textbf{+0.8} AP. 

Fig. \ref{fig:feature_map_comparison} visualises the shallow and deep features extracted from the backbones for the pre-trained, finetuned and our proposed methods. The noisy feature maps extracted from the pre-trained method clearly demonstrate that feature noise, caused by low-light conditions, confuse systems leading to poor performance. Upon finetuning the pre-trained model, we find that the feature extractor exhibits an excessive degree of denoising while our wNLBs effectively denoise the signal at multiple scales while preserving essential features, which results in a higher-quality semantic output.

Fig. \ref{fig:baseline_visual_comparison} illustrates the efficacy of our wNLB method in enhancing features, while also indicating areas for further refinement. Generally, the methods exhibit proficiency in segmenting larger objects but encounter challenges with smaller ones, likely attributed to injected noise. The \textit{easy} example case in Fig. \ref{fig:baseline_visual_comparison} had one large object of a cat and all models were able to accurately generate an instance mask for it.
In the \textit{medium} case, the pre-trained Mask R-CNN method failed significantly and falsely labelled many pixels. The finetuned Mask R-CNN performed better although it still suffered with accuracy when segmenting the tennis racket, while our method did not have this issue. In the \textit{difficult} case, none of the cows in the distance were detected by the methods.

Despite the marginal gains in AP seen in Table \ref{tab:methods_comparison}, we would like to highlight the qualitative improvements, seen in \ref{fig:baseline_visual_comparison}, especially for the medium case. We believe that it is important to observe the performance of the methods both quantitatively and qualitatively.

\begin{figure}[t]
\centering
    \includegraphics[width=\linewidth]{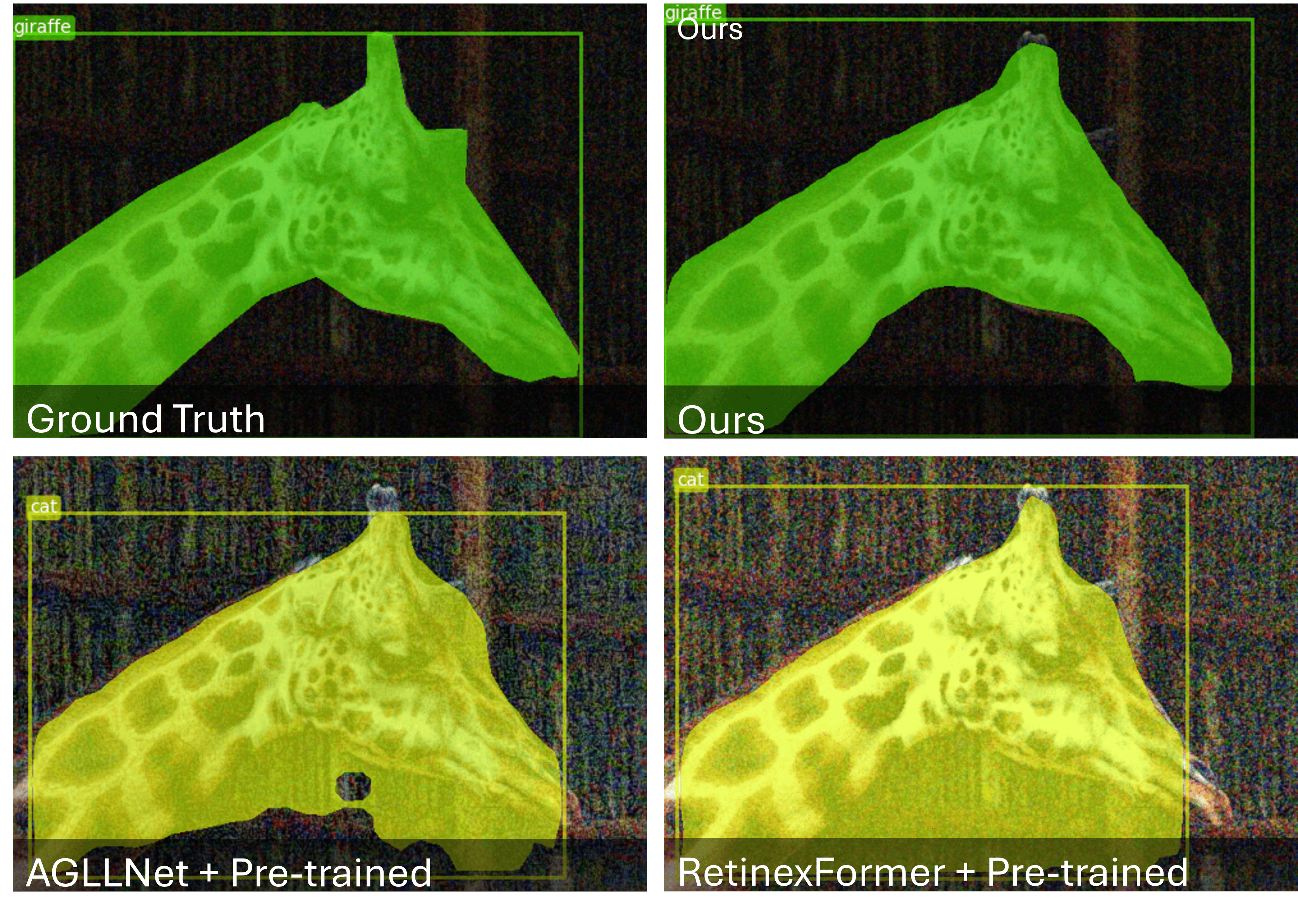}
    \caption{Visual comparison of our proposed method against two-stage methods (enhanced first then passed through a pre-trained model) using Mask R-CNN \cite{he2017maskrcnn} as the detector.}
    \label{fig:llie_visual_comparison}
    \vspace{-1.5em}
\end{figure}

\subsection{Two-stage experiments}
\label{ssec:preprocessing_experiments}
We also compared our method with the two-stage approaches, where the pre-processing employed popular LLIE methods, EnlightenGAN \cite{jiang2021enlightengan}, ZeroDCE++ \cite{li2021zerodce}, AGLLNet \cite{lv2021agllnet}, and RetinexFormer \cite{cai2023retinexformer} using off-the-shelf weights provided by the authors. 

It can be seen from Table \ref{tab:enhance_methods_comparison} that these approaches underperformed compared to using a pre-trained model, primarily because the contrast-enhanced images were impacted by amplified noise (see Fig. \ref{fig:llie_visual_comparison}), even with the transformer-based method RetinexFormer which attempted to reduce the noise amplification.
In the example shown, the two-stage methods made incorrect predictions for the giraffe as the amplified noise confused the model.
We found that using AGLLNet for the pre-processing step gave the best results for the two-stage methods. This is likely due to the fact we employed their proposed synthetic low-light pipeline for our synthetic low-light dataset.

\subsection{Qualitative assessment on a real low-light dataset}
\label{ssec:real_lowlight_experiments}
\begin{figure}[t]
\centering
    \includegraphics[width=\linewidth]{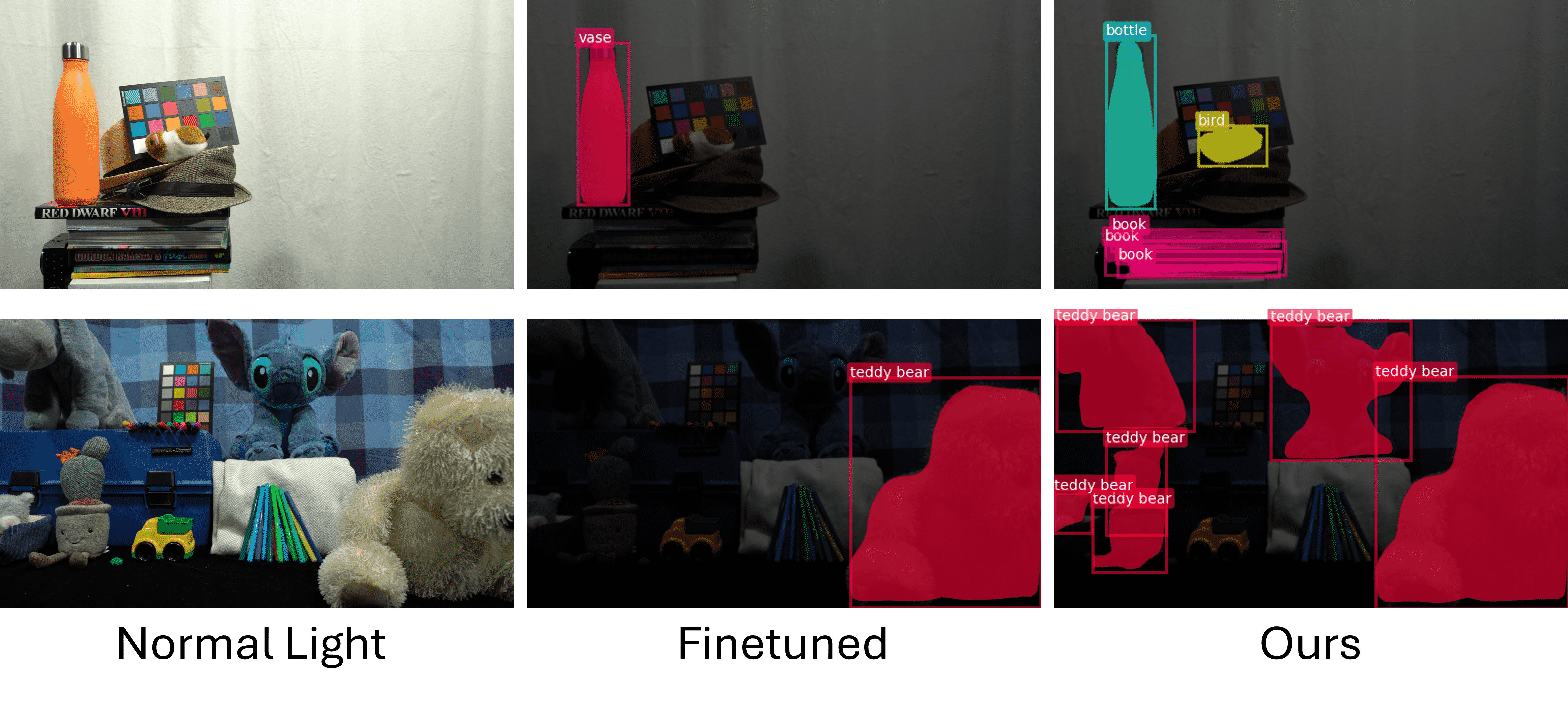}
    \caption{Visual comparison of our proposed method against the finetuned method using the Mask R-CNN \cite{he2017maskrcnn} architecture on real low-light data from the BVI-RLV Video dataset \cite{lin2024bvi}.}
    \label{fig:real_visual_comparison}
    \vspace{-1em}
\end{figure}

To verify the effectiveness of our method under real low-light conditions, we ran inference on video frames from the BVI-RLV Video Dataset \cite{lin2024bvi}, where the low-light videos have 10\% of the brightness to the normal-light counterparts. We compared the subjective performance of our wNLB method with the finetuned method, due to a lack of pixel-wise annotations. As shown in Fig. \ref{fig:real_visual_comparison}, our wNLB method segments more objects and classifies them more accurately than the finetuned model. This confirms that our method can also be applied to instance segmentation tasks under real low-light conditions.


\subsection{Non-local operation ablation study}
\label{ssec:nlm_experiments}

\begin{table}[hbt!]
\vspace{-0.5em}
    \caption{Ablation study of different NL operations for NL block with weights $w_1 = 0$, $w_2 = 0$, $w_3 = 0.5$, $w_4 = 0$}
    \centering
    \begin{tabular}{lccc}
        \toprule
        Method     &  AP & AP$_{50}$ & AP$_{75}$  \\
        \midrule
        Dot Product        & 16.1 & 29.3 & 15.7 \\
        Gaussian           & 16.0 & 29.1 & 15.7 \\
        Embedded Gaussian  & \textbf{16.6} & \textbf{30.2} & \textbf{16.4} \\
        \bottomrule
    \end{tabular}
    \label{tab:nlm_comparison}
\end{table}

We compared our approach with using different NL operations in order to determine which was the most effective denoising algorithm. As seen in Table \ref{tab:nlm_comparison}, they all show similar performances, but the embedded Gaussian form shows some improvement compared to the other two. The embedded Gaussian form likely had a higher performance due to its self-attention properties, as described by Wang et al. \cite{wang2018nonlocalnn}.

\section{Conclusion}
\label{sec:conclusion}

This paper introduces a framework for low-light instance segmentation without the need for pre-processing. We enhance popular instance segmentation models by proposing ``plug-and-play'' weighted non-local blocks (wNLB) for feature denoising. Experimental results show that our method improves the average precision by upto \textbf{+10.0} over vanilla models pretrained on normal-light images. In particular, our wNLB improves performances for smaller objects obscured by noise that would previously be undetected. Our approach is applicable to various low-light computer vision tasks, including object detection and semantic segmentation, by integrating wNLBs into the backbones of other architectures. Future work should address performance on even smaller objects, improve other aspects such as the mask head for more precise masks, and evaluate our method's efficacy against real low-light datasets with precise pixel-wise annotations.

\bibliographystyle{IEEEtran}
\bibliography{refs}

\begin{thebibliography}{10}
\providecommand{\url}[1]{#1}
\csname url@samestyle\endcsname
\providecommand{\newblock}{\relax}
\providecommand{\bibinfo}[2]{#2}
\providecommand{\BIBentrySTDinterwordspacing}{\spaceskip=0pt\relax}
\providecommand{\BIBentryALTinterwordstretchfactor}{4}
\providecommand{\BIBentryALTinterwordspacing}{\spaceskip=\fontdimen2\font plus
\BIBentryALTinterwordstretchfactor\fontdimen3\font minus \fontdimen4\font\relax}
\providecommand{\BIBforeignlanguage}[2]{{%
\expandafter\ifx\csname l@#1\endcsname\relax
\typeout{** WARNING: IEEEtran.bst: No hyphenation pattern has been}%
\typeout{** loaded for the language `#1'. Using the pattern for}%
\typeout{** the default language instead.}%
\else
\language=\csname l@#1\endcsname
\fi
#2}}
\providecommand{\BIBdecl}{\relax}
\BIBdecl

\bibitem{lin2014coco}
T.-Y. Lin, M.~Maire, S.~Belongie, J.~Hays, P.~Perona, D.~Ramanan, P.~Doll{\'a}r, and C.~L. Zitnick, ``Microsoft coco: Common objects in context,'' in \emph{European Conference on Computer Vision}, 2014, pp. 740--755.

\bibitem{zhou2017ade20k}
B.~Zhou, H.~Zhao, X.~Puig, S.~Fidler, A.~Barriuso, and A.~Torralba, ``Scene parsing through ade20k dataset,'' in \emph{Proceedings of the IEEE Conference on Computer Vision and Pattern Recognition}, 2017.

\bibitem{Everingham15pascalvoc}
M.~Everingham, S.~M.~A. Eslami, L.~Van~Gool, C.~K.~I. Williams, J.~Winn, and A.~Zisserman, ``The pascal visual object classes challenge: A retrospective,'' \emph{International Journal of Computer Vision}, vol. 111, no.~1, pp. 98--136, Jan. 2015.

\bibitem{cordts2016cityscapes}
M.~Cordts, M.~Omran, S.~Ramos, T.~Rehfeld, M.~Enzweiler \emph{et~al.}, ``The cityscapes dataset for semantic urban scene understanding,'' in \emph{Proceedings of the IEEE Conference on Computer Vision and Pattern Recognition (CVPR)}, June 2016.

\bibitem{he2017maskrcnn}
K.~He, G.~Gkioxari, P.~Dollár, and R.~Girshick, ``Mask r-cnn,'' in \emph{2017 IEEE International Conference on Computer Vision (ICCV)}, 2017, pp. 2980--2988.

\bibitem{jiang2021enlightengan}
Y.~Jiang, X.~Gong, D.~Liu, Y.~Cheng, C.~Fang, X.~Shen, J.~Yang, P.~Zhou, and Z.~Wang, ``Enlightengan: Deep light enhancement without paired supervision,'' \emph{IEEE Transactions on Image Processing}, vol.~30, pp. 2340--2349, 2021.

\bibitem{li2021zerodce}
C.~Li, C.~G. Guo, and C.~C. Loy, ``Learning to enhance low-light image via zero-reference deep curve estimation,'' in \emph{IEEE Transactions on Pattern Analysis and Machine Intelligence}, 2021.

\bibitem{lv2021agllnet}
F.~Lv, Y.~Li, and F.~Lu, ``Attention guided low-light image enhancement with a large scale low-light simulation dataset,'' \emph{International Journal of Computer Vision}, vol. 129, no.~7, pp. 2175--2193, 2021.

\bibitem{cai2023retinexformer}
Y.~Cai, H.~Bian, J.~Lin, H.~Wang, R.~Timofte, and Y.~Zhang, ``Retinexformer: One-stage retinex-based transformer for low-light image enhancement,'' in \emph{Proceedings of the IEEE/CVF International Conference on Computer Vision (ICCV)}, October 2023, pp. 12\,504--12\,513.

\bibitem{2018exdark}
Y.~P. Loh and C.~S. Chan, ``Getting to know low-light images with the exclusively dark dataset,'' \emph{Computer Vision and Image Understanding}, vol. 178, pp. 30--42, 2019.

\bibitem{2020poorvisdata}
W.~Yang, Y.~Yuan, W.~Ren, J.~Liu, W.~J. Scheirer \emph{et~al.}, ``Advancing image understanding in poor visibility environments: A collective benchmark study,'' \emph{IEEE Transactions on Image Processing}, vol.~29, pp. 5737--5752, 2020.

\bibitem{2023lis}
L.~Chen, Y.~Fu, K.~Wei, D.~Zheng, and F.~Heide, ``Instance segmentation in the dark,'' \emph{International Journal of Computer Vision}, vol. 131, no.~8, pp. 2198--2218, 2023.

\bibitem{Yi:Comprehensive:2024}
A.~Yi and N.~Anantrasirichai, ``A comprehensive study of object tracking in low-light environments,'' \emph{arXiv:2312.16250}, 2024.

\bibitem{wang2018nonlocalnn}
X.~Wang, R.~Girshick, A.~Gupta, and K.~He, ``Non-local neural networks,'' in \emph{Proceedings of the IEEE Conference on Computer Vision and Pattern Recognition (CVPR)}, June 2018.

\bibitem{ren2015fasterrcnn}
S.~Ren, K.~He, R.~Girshick, and J.~Sun, ``Faster r-cnn: Towards real-time object detection with region proposal networks,'' in \emph{Advances in Neural Information Processing Systems}, vol.~28, 2015.

\bibitem{vaswani2017attention}
A.~Vaswani, N.~Shazeer, N.~Parmar, J.~Uszkoreit, L.~Jones \emph{et~al.}, ``Attention is all you need,'' in \emph{Advances in Neural Information Processing Systems}, vol.~30, 2017.

\bibitem{carion2020detr}
N.~Carion, F.~Massa, G.~Synnaeve, N.~Usunier, A.~Kirillov, and S.~Zagoruyko, ``End-to-end object detection with transformers,'' in \emph{European conference on computer vision}.\hskip 1em plus 0.5em minus 0.4em\relax Springer, 2020, pp. 213--229.

\bibitem{jocher2023yolo}
\BIBentryALTinterwordspacing
G.~Jocher, A.~Chaurasia, and J.~Qiu, ``{Ultralytics YOLO},'' Jan. 2023. [Online]. Available: \url{https://github.com/ultralytics/ultralytics}
\BIBentrySTDinterwordspacing

\bibitem{wang2020solov2}
X.~Wang, R.~Zhang, T.~Kong, L.~Li, and C.~Shen, ``{SOLOv2}: Dynamic and fast instance segmentation,'' in \emph{Proc. Advances in Neural Information Processing Systems (NeurIPS)}, 2020.

\bibitem{pizer1987histeq}
\BIBentryALTinterwordspacing
S.~M. Pizer, E.~P. Amburn, J.~D. Austin, R.~Cromartie, A.~Geselowitz \emph{et~al.}, ``Adaptive histogram equalization and its variations,'' \emph{Computer Vision, Graphics, and Image Processing}, vol.~39, no.~3, pp. 355--368, 1987. [Online]. Available: \url{https://www.sciencedirect.com/science/article/pii/S0734189X8780186X}
\BIBentrySTDinterwordspacing

\bibitem{land1977retinex}
E.~H. Land, ``The retinex theory of color vision,'' \emph{Scientific american}, vol. 237, no.~6, pp. 108--129, 1977.

\bibitem{lore2017llnet}
K.~G. Lore, A.~Akintayo, and S.~Sarkar, ``Llnet: A deep autoencoder approach to natural low-light image enhancement,'' \emph{Pattern Recognition}, vol.~61, pp. 650--662, 2017.

\bibitem{Chen2018RetinexNet}
C.~Wei, W.~Wang, W.~Yang, and J.~Liu, ``Deep retinex decomposition for low-light enhancement,'' in \emph{British Machine Vision Conference}, 2018.

\bibitem{zhang2019kindling}
Y.~Zhang, J.~Zhang, and X.~Guo, ``Kindling the darkness: A practical low-light image enhancer,'' in \emph{Proceedings of the 27th ACM International Conference on Multimedia}, 2019, pp. 1632--1640.

\bibitem{2021nightcity}
X.~Tan, K.~Xu, Y.~Cao, Y.~Zhang, L.~Ma, and R.~W.~H. Lau, ``Night-time scene parsing with a large real dataset,'' \emph{IEEE Transactions on Image Processing}, vol.~30, pp. 9085--9098, 2021.

\bibitem{2020bdd100k}
F.~Yu, H.~Chen, X.~Wang, W.~Xian, Y.~Chen, F.~Liu, V.~Madhavan, and T.~Darrell, ``Bdd100k: A diverse driving dataset for heterogeneous multitask learning,'' in \emph{IEEE/CVF Conference on Computer Vision and Pattern Recognition (CVPR)}, June 2020.

\bibitem{buades2005nlm}
A.~Buades, B.~Coll, and J.-M. Morel, ``A non-local algorithm for image denoising,'' in \emph{IEEE Computer Society Conference on Computer Vision and Pattern Recognition (CVPR'05)}, vol.~2, 2005, pp. 60--65.

\bibitem{tomasi1998bilateral}
C.~Tomasi and R.~Manduchi, ``Bilateral filtering for gray and color images,'' in \emph{Sixth International Conference on Computer Vision (IEEE Cat. No.98CH36271)}, 1998, pp. 839--846.

\bibitem{xie2019nlmdenoise}
C.~Xie, Y.~Wu, L.~van~der Maaten, A.~L. Yuille, and K.~He, ``Feature denoising for improving adversarial robustness,'' in \emph{The IEEE Conference on Computer Vision and Pattern Recognition (CVPR)}, June 2019.

\bibitem{he2016resnet}
K.~He, X.~Zhang, S.~Ren, and J.~Sun, ``Deep residual learning for image recognition,'' in \emph{2016 IEEE Conference on Computer Vision and Pattern Recognition (CVPR)}, 2016, pp. 770--778.

\bibitem{lin2017fpn}
T.-Y. Lin, P.~Dollár, R.~Girshick, K.~He, B.~Hariharan, and S.~Belongie, ``Feature pyramid networks for object detection,'' in \emph{2017 IEEE Conference on Computer Vision and Pattern Recognition (CVPR)}, 2017, pp. 936--944.

\bibitem{lin2024bvi}
R.~Lin, N.~Anantrasirichai, G.~Huang, J.~Lin, Q.~Sun, A.~Malyugina, and D.~Bull, ``{BVI-RLV}: A fully registered dataset and benchmarks for low-light video enhancement,'' \emph{arXiv preprint arXiv:2407.03535}, 2024.

\end{thebibliography}

\end{document}